\documentclass[conference]{IEEEtran}
\IEEEoverridecommandlockouts
\usepackage{cite}
\usepackage{amsmath,amssymb,amsfonts}
\usepackage{algorithmic}
\usepackage{graphicx}
\usepackage{textcomp}
\usepackage{xcolor}
\usepackage{subcaption}
\usepackage{multirow}
\usepackage{booktabs}
\usepackage{hyperref}
\def\BibTeX{{\rm B\kern-.05em{\sc i\kern-.025em b}\kern-.08em
    T\kern-.1667em\lower.7ex\hbox{E}\kern-.125emX}}
\begin{document}

\title{MAD: Self-Supervised Masked Anomaly Detection Task for Multivariate Time Series \\
}

\author{\IEEEauthorblockN{Yiwei Fu}
	\IEEEauthorblockA{\textit{GE Research} \\
		Niskayuna, NY, USA \\
		yiwei.fu@ge.com}
	\and
	\IEEEauthorblockN{Feng Xue}
	\IEEEauthorblockA{\textit{GE Research} \\
		Niskayuna, NY, USA \\
		xue@ge.com}
}
\maketitle

\begin{abstract}
In this paper, we introduce \textbf{M}asked \textbf{A}nomaly \textbf{D}etection (\textbf{MAD}),
a general self-supervised learning task for multivariate time series anomaly detection.
With the increasing availability of sensor data from industrial systems, being able to detecting anomalies from 
streams of multivariate time series data is of significant importance. 
Given the scarcity of anomalies in real-world applications, the majority of literature has been focusing on modeling normality.
The learned normal representations can empower anomaly detection as the model has  learned to capture
certain  key underlying data regularities.
A typical formulation is to learn a predictive model, i.e., use a window of time series data to
predict future data values. 
In this paper, we propose an alternative self-supervised learning task.
By randomly masking a portion of the inputs and training a model to estimate them using the remaining ones, MAD is an improvement over the traditional left-to-right next step prediction (NSP) task.
Our experimental results demonstrate that MAD can achieve better anomaly detection rates over traditional NSP
approaches when using exactly the same neural network (NN) base models,
and can be modified to run as fast as NSP models during test time on the same hardware, thus making it an ideal upgrade for many existing NSP-based NN anomaly detection models.

\end{abstract}


\section{Introduction}
Anomaly detection has been widely studied and is of significant importance in
many application areas such as fraud detection, cyber security, and complex system health monitoring
\cite{chandola_anomaly_2009}. In recent years, deep learning has seen increasing adoptions in
anomaly detection 
\cite{chalapathy_deep_2019}. 
Of particular interest to this paper
is anomaly detection on industrial multivariate time series data. With the increasing number of sensors as well as 
cost-effective data
transmission and storage solutions, industrial systems (such as power plants, wind turbines, engines, etc.) generate large 
amounts of
time series data during their operations. It is important to monitor these systems for spotting abnormal behaviors,
which could lead to significant reliability consequences.

Anomalies, also referred to as outliers, are observations which deviate so much from the majority of all the other
ones. In the context of industrial time series data, systems are usually being operated under their designed normal operating conditions (NOCs),  and these system measurements partially capture of the dynamic
states governed by first principles and underlying control logic.
The core idea of developing anomaly detection is
to learn the spatial (across multiple system measurements and commands) and temporal relationships under NOC. In an abnormal situation, such relationships will not adhere to the learned representation, resulting in
deviations  from the normal operating patterns. The higher a deviation is, the more likely there is an anomaly.

In the process control and model-based fault detection community, a number of data driven approaches have been
used for anomaly detection of industrial time series data, for example, Principal Component Analysis (PCA),
Dynamic PCA
\cite{ku_disturbance_1995}, subspace aided approach \cite{ding_subspace_2009}. These traditional approaches take into account
some multivariate linear relationships, and to some extent temporal dependencies (in the cases of Dynamic PCA and subspace
aided approach). Another commonly-used approach is one-class SVM such as the one presented in~\cite{ma_time-series_2003}. In recent years, however, deep
learning has taken the center stage for multivariate anomaly detection \cite{
	hundman_detecting_2018, malhotra_lstm-based_2016, guo_multidimensional_2018, zong_deep_2018, su_robust_2019, xue_deep_2020}.

One of the most common deep learning tasks for modeling normal time series data is next step prediction (NSP), in which past observations within a temporal window are used to predict the future. A Recurrent Neural Network (RNN) such as
an LSTM~\cite{hochreiter_long_1997} is usually
used for such tasks, although recent work \cite{oord_wavenet_2016, bai_empirical_2018} demonstrated that
a causal convolutional
network might be a good alternative to RNNs in term of effectiveness and training efficiency. 
NSP models have the basic assumption that normal instances are temporally more predictable than anomalies~\cite{pang2020deep}. Based on the same assumption and partially inspired by BERT \cite{devlin_bert_2019}, it should also be true that masked instances anywhere in the temporal window, as opposed to only the last steps in the NSP task, are more predictable for normal data than abnormal ones.

In this paper,
we formally propose MAD (Masked Anomaly Detection), a self-supervised anomaly detection task for time series data representation learning, where models are trained to estimate masked values anywhere in a window of time series data.
This task will enable learning from bidirectional contexts (when the base model allows, such as
Transformer~\cite{vaswani_attention_2017}). In contrast, NSP models is limited to left-to-right unidirectional context.
Furthermore, even when the base model is inherently not bidirectional (such as LSTM), the MAD task is still valid as a task to reconstruct the masked values from previous inputs. 
It extends the NSP task in that a NSP task can be regarded as a MAD model with masking only the last steps in a 
sequence. MAD is also very flexible: it allows for the use of any neural network base models that can generate a 
sequence, and during test time it can run slower but more accurate (by masking all steps sequentially) or as fast and 
accurate as NSP models (by masking only the last step).

The contributions of this paper are as follows:

\begin{itemize}
	\item We proposed a self-supervised learning task, MAD, for time series anomaly detection. To our knowledge, this is
	the first attempt to use masked self-supervised learning for multivariate time series anomaly detection.
	Although masked language model (MLM) has been studied before, we extend this to industrial anomaly detection scenarios where the data is continuous and multivariate.
	\item Our experiments demonstrated the superior performance of this learning task over the
	traditional NSP task in anomaly detection. We also proposed two inference modes (MAD vs. Fast-MAD) for
	anomaly detection, allowing trading off a small accuracy loss from MAD for faster inference in Fast-MAD. The latter 
	performs an NSP-like inference by masking only the last steps, but with better or similar accuracy.
	\item Unlike BERT where the underlying model architecture is fixed, we show that MAD can accommodate any
	base model 
	that generates sequence outputs, and improve all of them over the NSP task. This flexibility has significant 
	implications for real-world applications: existing NSP models can be improved by 
	simply switching to our proposed MAD framework. Since the base models are the same, they should be able to run in the same hardware for better performance.
\end{itemize}

We emphasize that the goal of this paper is not to find a set of hyperparameters for one particular model that performs better on some benchmark datasets over other models. 
Instead, we are focusing on a more general setting and trying to present a new learning task that is widely applicable regardless of the choice of base models and their hyperparameters. 
We also acknowledge that other learning tasks could be formulated (e.g., end-to-end anomaly score learning, direct classification with abundant anomaly data, etc.), 
but for fair comparison they are not presented here because they generally cannot use the same data, base models, or hyperparameters.

\section{Related Work}
\subsection{Masked Language Models}
BERT~\cite{devlin_bert_2019}
is a popular language representation learning model. Based on the concept of Masked Language Model (MLM), BERT is
designed to pre-train deep bidirectional representations from unlabeled text. The idea that a representation which is able to learn the
context around a word rather than just before the word is able to better capture its meaning provides
inspirations for us that similar mechanisms could potentially also be useful for time series anomaly detection.

Following BERT's success, there is an explosion of recent work that either tries to improve BERT itself or apply BERT
to some other domains. For example, RoBERTa~\cite{liu2019roberta} tries to improve BERT by training longer and
with bigger batches, removing the next sentence prediction objective (thus leaving only the MLM objective), using
longer sequences, and changing masking patterns dynamically. ALBERT~\cite{lan2019albert} is a lightweight
version of BERT with fewer parameters and faster training. XLNet~\cite{yang2019xlnet} leverages both autoregressive
language models and contextual-based BERT pretraining while attempting to avoid their respective limitations. On
the one hand, XLNet uses bidirectional context like BERT; on the other hand, it works as an autoregressive language model and
does not rely on data masking.

Different from these MLMs, our proposed MAD framework is focused on multivariate time series data. The continuous
nature of  industrial data (mostly from sensor measurements) means that we cannot have a discrete mask token, and
a new masking procedure needs to be examined. We also do not limit the base model to Transformer only.

\subsection{Self-Supervised Representation Learning}
Self-supervised learning enables supervised-like learning without a labeled dataset by creating artificial ``labels'' for
free from the data itself. This can be achieved by formulating the learning task as predicting a subset of information
using the rest (e.g., predicting the future from the past, past from present, or a part of the input from the rest, etc.).
BERT~\cite{devlin_bert_2019} also falls into this category. Usually, the self-supervised task (or the pre-train task) is a
means to an end: it learns a useful representation that can be beneficial to some downstream tasks.

Various self-supervised learning approaches have been proposed for images. Pathak et al.~\cite{pathak2016context}
use inpainting to fill in the missing parts in an image. Doersch et al.~\cite{doersch2015unsupervised} formulated the
self-supervised task as learning the relative position between two random patches in an image. Noroozi \& Favaro
extended this to jigsaw puzzles~\cite{noroozi2016unsupervised}. Other methods uses various different tasks, for
example, colorization~\cite{zhang2016colorful,larsson2016learning}, geometric
distortion~\cite{dosovitskiy2015discriminative,gidaris2018unsupervised}, noise as
targets~\cite{bojanowski2017unsupervised}, clustering~\cite{caron2018deep},
contrast learning~\cite{he2020momentum, chen2020simple}. Many of these pretext tasks are easily constructed on
images and the learning
can be automatically done on the dataset by requiring the network to recover the modified information in these
images.

There has also been some work on videos, because spatiotemporal data are a natural fit for the self-supervised learning scheme,
although in general they are more challenging than images. LSTM and its variants for future prediction are the most
common models~\cite{srivastava2015unsupervised,fu2020spatiotemporal}. Wang et
al.~\cite{wang2015unsupervised} used visual tracking to learn visual representations. Misra et
al.~\cite{misra2016shuffle} formulated the pretext task as a sequential verification task for determining whether a
sequence of frames from a video is in the correct temporal order. Vondrick et al.~\cite{vondrick2018tracking}
extend image colorization to videos and find that the model learns to track visual regions. Ali et
al.~\cite{ali2020self} used a frame permutation prediction task for visual anomaly detection.

\subsection{Transformers for Time Series}
Transformer~\cite{vaswani_attention_2017} has seen tremendous success in replacing traditional RNNs for sequence modeling, in not only the language domain but also many others.

Specifically, Zerveas et al.~\cite{zerveas2020transformer} used Transformer for unsupervised multivariate time series representation learning, but the authors focus on regression and classification. This approach was shown to outperform supervised methods in benchmark datasets. Wu et al.~\cite{wu2020deep} developed a Transformer model for forecasting influenza prevalence and showed superior performance over previous models. Li et al.~\cite{li2019enhancing} used a variant of Transformer that has a lower memory cost for long-sequence univariate time series forecasting.

There exists some other work that utilizes Transformers for anomaly detection. Guo et al.~\cite{guo2021logbert}
used BERT for detecting anomalous events in online computer systems by learning the patterns of normal log
sequences. This work is closer to language than time series because logs are basically language data.
Meng et al.~\cite{meng2019spacecraft} used a transformer-based architecture for spacecraft anomaly detection.
The authors developed a specific masking scheme where the front and end parts of a sequence are used as inputs to
the model, and the middle part is masked. In contrast, our proposed MAD framework is more general, can incorporate
different base models, and is not limited to a specific dataset and problem setup.

\section{Masked Anomaly Detection}~\label{sec:problem_formulation}
In general, anomaly detection with neural networks can be categorized into three paradigms: deep learning for
feature extraction, learning feature representation of normality, and end-to-end anomaly score
learning~\cite{pang2020deep}. In a typical industrial setting, normal operation data are usually abundant while the
number of faulty cases is often very small (if there are any). Therefore, the paradigm of modeling normality
is often used: it learns a representation of data by using an objective function that is not directly
measuring an anomaly score, but the learned normal representation can still be useful in an anomaly detection setting since
they capture some underlying properties of the normal data.

\begin{figure}[t]
	\centering
	\begin{subfigure}{1.0\linewidth}
		\centering
		\includegraphics[width=0.85\linewidth]{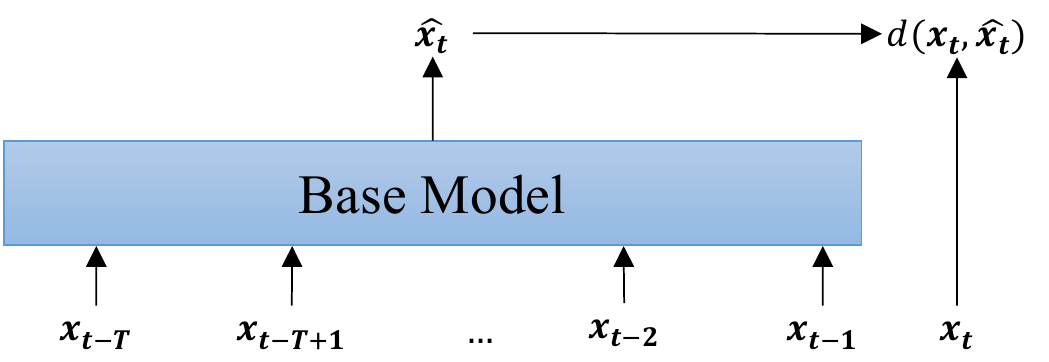}
		\caption{Traditional next step prediction (NSP) formulation for anomaly detection: Given a sequence $[\mathbf{x_{t-T},...,x_t}]$, the first $T$ inputs are used to predict $\hat{x_t}$, and then the distance between $\hat{\mathbf{x_t}}$ and $\mathbf{x_t}$ is calculated.}
		\label{fig:formu_ar}
	\end{subfigure}
	\begin{subfigure}{1.0\linewidth}
		\centering
		\includegraphics[width=0.95\linewidth]{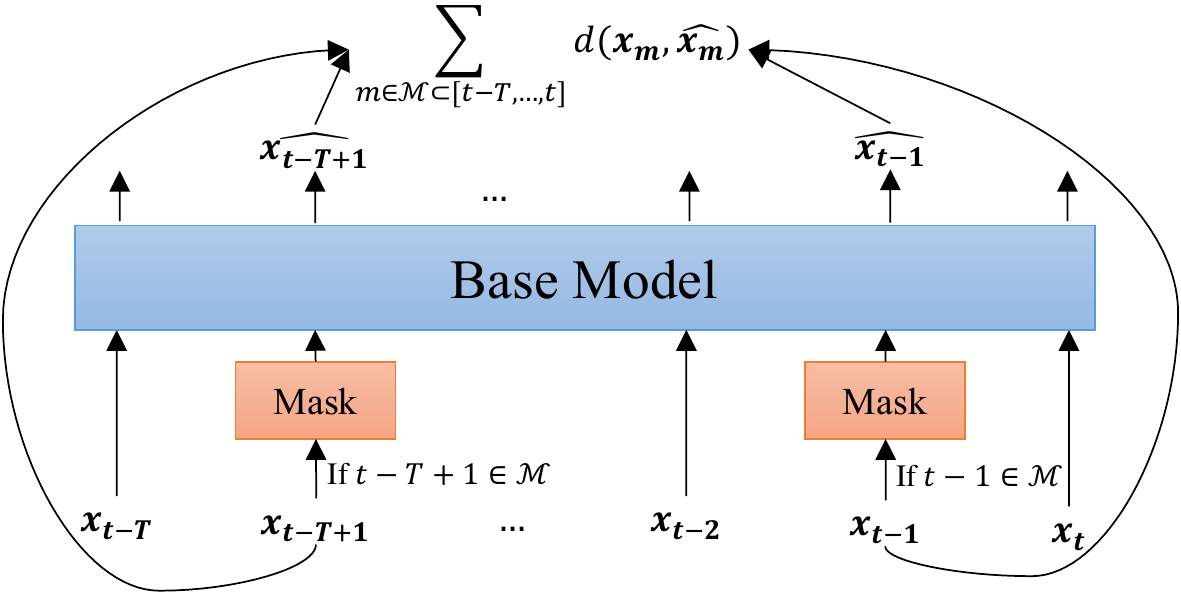}
		\caption{MAD, our proposed self-supervised anomaly detection task: Given a sequence $[\mathbf{x_{t-T},...,x_t}]$, some  inputs $\mathbf{x_m}$ are masked if $m\in\mathcal{M}$, where $\mathcal{M}$ is a randomly chosen subset of $[t-T,...,t]$. The model learns to estimate the masked values and calculates the distance between $\hat{\mathbf{x_m}}$ and $\mathbf{x_m}$ for all $m\in\mathcal{M}$.}
		\label{fig:formu_ss}
	\end{subfigure}
	\caption{Comparison between traditional NSP anomaly detection formulation and our proposed MAD task.}
	\label{fig:formulation}
\end{figure}

Traditionally, anomaly detection problems in an industrial setting often uses a next step prediction (NSP) task 
as shown in Figure~\ref{fig:formu_ar}. Let $\mathbf{x_t}\in \mathbb{R}^n$ be the multivariate sample of dimension $n$ at time $t$, and denote the $j$-th dimension at time $t$ as $x^j_t$ (i.e., $\mathbf{x_t}=[x_t^1, x_t^2,...,x_t^n]$), the NSP approach is trying to estimate $\mathbf{x_t}$ from all observations up to time $t-1$.
In practice, a window of length $T$ is often used as the inputs to the model instead of all samples prior to time $t$. This window length can be adjusted to different applications and datasets. The distance metric $d$ can be chosen as the Euclidean distance, corresponding to a mean squared error (MSE) loss during training as in Equation~\ref{eq:mse_ar}:
\begin{equation}\label{eq:mse_ar}
	L_{mse} = \frac{1}{n}\sum_{j=1}^n (\hat{x_t^j}-x_t^j)^2
\end{equation}
This distance metric measures the deviation of a sample from what it should have been under normal operating conditions. Therefore, a sample whose deviation is above a defined threshold can be regarded as anomalous. 

Given a sequence of time series data $\mathbf{[\mathbf{x_{t-T},...,x_t}]}$, the NSP approach only uses the data in one direction. 
%
However, there is no need to constrain the model to only learn normality unidirectionally. Instead, partially inspired by the
success of BERT~\cite{devlin_bert_2019} in language modeling tasks, we propose a general self-supervised learning
task for anomaly detection, MAD. Under this formulation, the task is to model the normal
data $[\mathbf{x_{t-T},...,x_t}]$ by randomly masking a portion of the inputs and training a model to
estimate those masked samples. 
Note that we are using the same window here as the NSP task.
By formulating the learning task this way, there are two major benefits:
\begin{enumerate}
	\item The model is able to utilize the training data more efficiently because the same sequence can be masked in different ways, which effectively creates more training data.
	\item The model is more comprehensive than unidirectional NSP models, because the latter one can be viewed as a special case of when only the last sample is masked.
\end{enumerate}

More formally, our proposed MAD formulation for anomaly detection is shown in Figure~\ref{fig:formu_ss}. Given a sequence $[\mathbf{x_{t-T},...,x_t}]$, some  inputs $\mathbf{x_m}$ are masked if $m\in\mathcal{M}$, where $\mathcal{M}$ is a randomly chosen subset of $[t-T,...,t]$ indicating the indices to be masked. 
Since too little masking makes the model expensive to train and too much masking would give not enough context, we follow the BERT~\cite{devlin_bert_2019} setup of masking out $15\%$ of the samples in all our experiments (i.e., $|\mathcal{M}|=(T+1)*15\%$ for a sequence). Different from the language models used in BERT where input tokens are discrete, industrial time series are often continuous. Therefore, we do not have a single discrete mask token. Instead, we replace all masked samples with random values within the input ranges. Then the masked sequence would be provided to the model to predict those masked values, and a distance metrics $d(\cdot,\cdot)$ is used to calculate the deviations between the predicted values and real values. During training, $\sum_{m\in\mathcal{M}} d(\mathbf{x_m,\hat{x_m}})$ can be used to calculate the loss. For example, the MSE loss for this formulation is given in Equation~\ref{eq:mse_ss}:
\begin{equation}\label{eq:mse_ss}
	L_{mse} = \frac{1}{|\mathcal{M}|}\frac{1}{n}\sum_{m\in\mathcal{M}}\sum_{j=1}^n (\hat{x_m^j}-x_m^j)^2
\end{equation}
where $n$ is the spatial dimension of the time series data at a certain time step.
It should be noted that a univariate time series can be think of as a special case for this formulation where $n=1$.

\begin{figure}[t]
	\centering
	\includegraphics[width=1.0\linewidth]{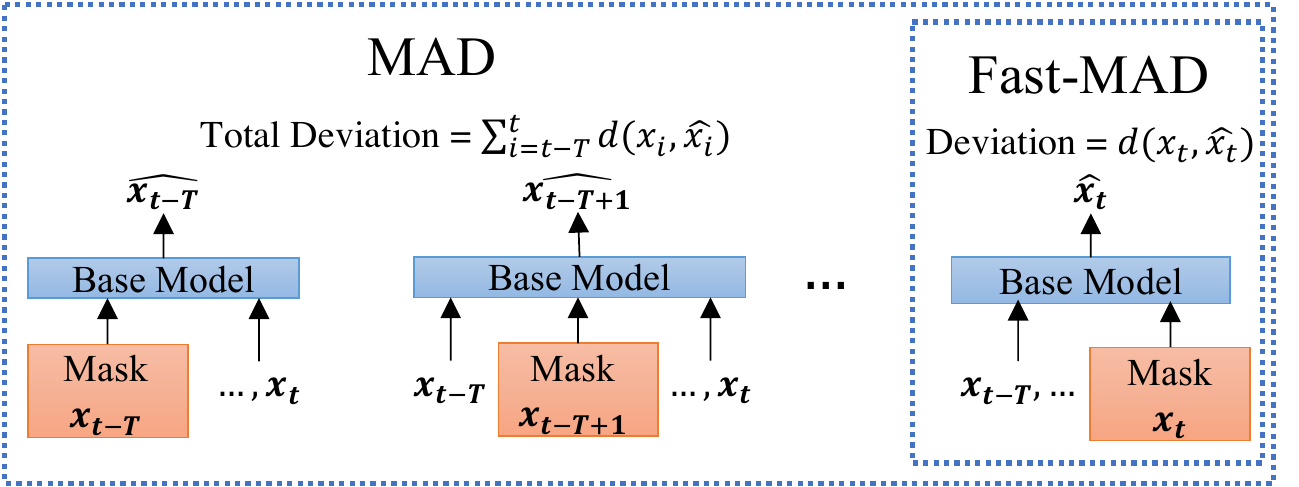}
	\caption{Anomaly detection after a MAD model is trained. Given a new sequence $[\mathbf{x_{t-T},...,x_t}]$, we can mask each sample $\mathbf{x_i}$ separately and estimate the normal values $\mathbf{\hat{x_i}}$, then calculate the total deviation for anomaly detection. Alternatively, we can mask only the last step for faster inference, which we call Fast-MAD.}
	\label{fig:inference}
\end{figure}

After a MAD model is trained on normal data, it can be used in anomaly detection. Fundamentally, the model learns a representation
of the normal data by being able to fill in the blanks of those masked inputs, thus when an anomaly happens it
would predict a different value given a mask. 
In Figure~\ref{fig:inference} we show the anomaly detection evaluation used in later sections for multivariate time series data. During the anomaly detection phase, we can mask every sample separately and calculate the total
deviation for the entire sequence. If the deviation is above a defined threshold, then this sequence can be regarded as
anomalous. Contrasting to the anomaly detection of traditional NSP methods in Figure~\ref{fig:formu_ar} where the
deviation is calculated only with the last sample $d(\mathbf{x_t,\hat{x_t}})$, our MAD task calculates the following:
\begin{equation}~\label{eq:total_dev}
	\sum_{i=t-T}^t d(\mathbf{x_i,\hat{x_i}})
\end{equation}
thus, it is able to better capture anomalies in time series data by looking at more than just the last step. Furthermore, in this way MAD models use masks
consistently between training and testing, unlike the discrepancy in BERT between pretraining and task specific fine
tuning where the artificial mask symbol is only introduced in the pretrain phase but not used in the fine tune phase.

One potential disadvantage of MAD is that it has a higher time complexity during inference. This 
can
be seen from Equation~\ref{eq:total_dev}: because we have to mask each time step separately and sum them up, MAD makes
$T$ times more computations than the traditional NSP formulation. In situations where 
speed or early alarm is crucial during inference, we can make use of the same MAD-trained model by
only masking the last time step, and then calculate the distance $d(\mathbf{x_t,\hat{x_t}})$. We call this
\textit{Fast-MAD} as shown in Figure~\ref{fig:inference}. In fact, this is very similar to how traditional NSP models perform a test, both in terms of the
metrics calculated and time complexity. Theoretically, Fast-MAD should run fast as traditional NSP during test time, and we will verify this in the experiments section.

Therefore, after a MAD model is trained, it has the flexibility to make 
a prediction as fast as NSP models by only masking the last time step, or slower but more accurate by masking each step
(or anywhere in between). This flexibility does not exist in traditional NSP approaches: 
if we want to predict  different-length outputs, different models have to be trained. In terms of space complexity and NN model size, since they are using the same base model, both MAD and NSP are basically equivalent. It means that for any real-world application that is already running NSP models, there is no reason that the same hardware could not handle MAD instead.

\section{Experimental Results}~\label{sec:exp}
In this section, the effectiveness of the proposed MAD task for anomaly detection is demonstrated on two case 
studies: the Tennessee Eastman Process (TEP)  dataset~\cite{downs1993plant} and the HIL-based Augmented ICS 
(HAI) security dataset~\cite{shin2020hai}.

For comparison, in both of these case studies, anomaly detection with both the traditional NSP formulation and our 
proposed MAD formulation (as in Figure~\ref{fig:formulation}) using various base models are tested. Base models and data inputs are kept the same across the two 
tasks. All the input variables from the datasets are scaled to a range of $[0,1]$ before they are passed to the 
models. 

The main objective of this paper is to compare the two anomaly detection tasks. 
Therefore, we present results from the same sets of hyperparameters for each base
model
across the two formulations. For the same reason, we also excluded more complicated post-processing methods 
for anomaly detection. Instead, we
deliberately kept the comparisons simple by using the same metrics (i.e., MSE) during the training and anomaly
detection phases to limit other contributing factors.


\subsection{Datasets}
\subsubsection{TEP Dataset}\footnote{TEP dataset can be downloaded at \url{https://dataverse.harvard.edu/dataset.xhtml?persistentId=doi:10.7910/DVN/6C3JR1}.}
The Tennessee Eastman process (TEP) is an industrial benchmark by the Eastman Chemical Company for process monitoring and control studies~\cite{downs1993plant}. It models a real industrial process computationally and is widely studied for anomaly detection algorithms~\cite{yin2012comparison, sun2020fault}.

The TEP is comprised of $4$ reactants, $2$ products, $1$ by-product and $1$ inert components denoted as A-H. These components undergo a chemical process enabled by 5 major units: a reactor where the reaction happens for the gas feed components (A, C, D and E) into liquid products (G and H), a condenser that  cools down the gas stream coming out of the reactor, a separator that separates gas and liquid components from the cooled product stream, a compressor that feeds the gas stream back into the reactor and a stripper that strips the two products from any unreacted feed components.

The TEP dataset~\cite{DVN/6C3JR1_2017} contains $52$ variables in total, $41$ of which are sensor measurements (XMEAS(1) - XMEAS(41)) and $11$ are manipulated variables ((XMV(1) - (XMV11)). Therefore, the multivariate samples $\mathbf{x_t}$ have a dimension of $52$, or $n=52$ following our previous notation. The dataset is divided into ``fault-free'' and ``faulty'' files. The former corresponds to the processes under normal operating conditions (NOC) while the latter contains 20 different simulated process faults. The fault-free training data consists of sequences of length $500$. The test data sequence length is $960$, with the first $160$ samples in the normal region, and faults are introduced for the next $800$ samples. For fault detection rate (FDR) calculations, we only considered the length $800$ faulty region:
\begin{equation}
	\mathrm{FDR=\frac{number \: of \: alarm \: samples \: after \: introducing \: fault}{total \: number \: of \: samples \: after \: introducing \: fault}}
\end{equation}
And the false alarm rate (FAR) is calculated as:
\begin{equation}
	\mathrm{FAR=\frac{number \: of \: alarm \: samples \: during \: NOC}{total \: number \: of \: samples \: during \: NOC}}
\end{equation}

For TEP data, we take a window of length $20$ as the inputs to the NSP models and length $1$ as the output. That is to
say, we use $[\mathbf{x_{t-20}}, \mathbf{x_{t-19}},...,\mathbf{x_{t-1}}]$ to estimate $\mathbf{x_{t}}$. For our
proposed MAD approach, the entire sequence, i.e., $[\mathbf{x_{t-20}}, \mathbf{x_{t-19}},...,\mathbf{x_{t}}]$, is
used for masked training and anomaly detection.

It should be noted that in this dataset, controllable faults (Fault $3,9,15$) have disturbances that can be dealt with by
the control system (thus ``controlable''), and therefore they would return to normal operating regions. In these circumstances, the FDR is not expected to be
significantly different from the FAR~\cite{sun2020fault}.

\subsubsection{HAI Dataset} \footnote{HAI dataset can be downloaded at \url{https://github.com/icsdataset/hai}.}
The HIL-based Augmented ICS (HAI) Security Dataset was collected from a realistic industrial control system (ICS) testbed augmented with a Hardware-In-the-Loop (HIL) simulator that emulates steam-turbine power generation and pumped-storage hydro power generation~\cite{shin2020hai}. Both normal and abnormal behaviors for ICS anomaly detection are included in the dataset, with the abnormal one collected based on various attack scenarios with the six control loops in three different types of industrial control devices.

The HAI testbed consists of four processes: Boiler Process (P1), Turbine Process (P2), Water-treatment Process (P3)
and HIL Simulation(P4). During normal operation, it is assumed that the operator operates the control facility in a
routine manner, while abnormal behaviors occur when some of the parameters are outside the  normal range or are in
unexpected states due to attacks, malfunctions, and failures. This study is conducted based on the $20.07$ version of
HAI dataset, which has a training and testing  dataset with $n=59$ process measurements to model. The data also
contains label information about whether there is an attack and where in the three processes.
There are a total of 177 hours of data in the training set and 123 hours of data in the test set.

Since we have already looked into different faulty cases in the TEP dataset, here for simplicity we only examine the
overall anomaly detection performance across all different attack locations (i.e., we treat all attacks as one ``abnormal'' class).  Following the TEP dataset formulation,
we also use a sequence of length $20$, i.e., $[\mathbf{x_{t-20}}, \mathbf{x_{t-19}},...,\mathbf{x_{t}}]$  at time $t$ for training and
anomaly detection.
Additionally, if any of the samples inside a sequence contains an attack, we label the entire sequence as
abnormal.

\subsection{Base Model Setups}
Here we introduce some base models used in the experiments for evaluating MAD vs NSP anomaly detection. 
Transformer~\cite{vaswani_attention_2017} and
its encoder-only variants are natural fits for MAD, because of their abilities to ``fill in the blanks'', i.e., to
reconstruct the masked inputs or to predict missing parts of the inputs from the entire sequence bidirectionally. But any model that supports seq2seq modeling can be used as a base model. Specifically, we also tested Long Short-Term Memory (LSTM) network~\cite{hochreiter_long_1997}, Temporal Convolutional Network (TCN)~\cite{bai_empirical_2018}, and FNet~\cite{lee2021fnet}. The mix of commonly-used  time series models and newer NLP models should provide good insights into the comparison between the two tasks. 

For each base model, their hyperparameters are set as follows:
\begin{itemize}
	\item \textbf{Transformer}~\cite{vaswani_attention_2017}: model dimension $128$, feed-forward dimension $512$, number of heads $8$, number of encoder \& decoder layers $6$, and dropout rate $0.1$.
	\item \textbf{Transformer-Encoder}~\cite{vaswani_attention_2017}: same as Transformer, but without decoders.
	\item \textbf{LSTM}~\cite{hochreiter_long_1997}: $2$ hidden layers each with dimension $50$.
	\item \textbf{TCN}~\cite{bai_empirical_2018}: $2$ hidden layers each with channel size $50$, convolutional kernel size $7$ and stride $1$, dilation factor is $2^i$ where $i$ is the $i$-th layer, and dropout rate $0.2$.
	\item \textbf{FNet}~\cite{lee2021fnet}: same as Transformer-Encoder, but using 2D FFT encoders with real and imaginary parts  concatenated (Lee-Thorp et al. in~\cite{lee2021fnet} used 1D FFT  encoders and discarded imaginary parts for language data). 
\end{itemize}

In all experiments, we trained on a GPU using an Adam optimizer~\cite{kingma_adam_2015} with $\beta_1=0.9$, $\beta_2=0.999$, $\epsilon=10^{-8}$ and a learning rate of $0.001$. Model batch size is set to be $1000$ and number of epochs is $1000$. The metrics for calculating loss during training and deviations during anomaly detection is MSE. For both training sets, 80\% of the data was used for training and 20\% for validation. For our MAD formulation, unless otherwise noted, we use a mask probability of $0.15$, and $100\%$ of the masked values are replaced with random uniform numbers drawn from the normalized data range.


\subsection{Results and Discussions}\label{sec:exp:results}
\begin{table*}[t]
	\centering
	\caption{TEP dataset fault detection results. The FAR for the NOC is set to be 5\%, and the FDR (in percentage) is
		listed for each of the 20 faults. We report the results using our proposed MAD task and the traditional NSP
		task. If for the same base model, there is a greater than 1\% differences for a certain fault, then the higher one
		is presented in bold.}
		\begin{tabular}{c|cc|cc|cc|cc|cc}
			\toprule
			Base Model & \multicolumn{2}{c|}{Transformer} & \multicolumn{2}{c|}{Transformer-Encoder} &  \multicolumn{2}{c|}{LSTM}  &\multicolumn{2}{c|}{TCN}  &\multicolumn{2}{c}{FNet}\\
			\midrule
			Fault \# & MAD & NSP & MAD & NSP & MAD & NSP & MAD & NSP & MAD & NSP \\
			\hline
			1 & 99.55 & 99.64 & 99.56 & 99.66 & 99.54 & 99.74 &  99.42 & 99.61 & 99.48 & 99.67 \\
			2 & 98.38 & 98.45 & 98.37 & 98.40 & 98.57 & 97.99 & 98.58 & 97.93  & 98.59 & 98.64 \\
			3 & 5.40 & 5.00 & 5.41 & 5.03 & 5.82 & 5.12 & 5.80 & 5.12  & 4.97 & 4.86 \\
			4 & 99.97 & 99.96 & 99.97 & 99.99 & 99.96 & 99.95 & 99.88 & 100.00  & 99.90 & 99.18 \\
			5 & \textbf{85.42 }& 28.86 & \textbf{80.00} & 25.74 & \textbf{86.21} & 20.63 & \textbf{55.15} & 26.46  & \textbf{56.70} & 25.82 \\
			6 & 99.97 & 100.00 & 99.97 & 100.00 & 99.92 & 100.00 & 99.91 & 100.00  & 99.92 & 100.00 \\
			7 & 99.99 & 100.00 & 99.99 & 100.00 & 99.99 & 100.00 & 99.95 & 100.00  & 99.98 & 100.00 \\
			8 & \textbf{97.57} & 96.43 & \textbf{97.52} & 95.80 & \textbf{97.57} & 93.42 & \textbf{97.49} & 94.68  & 97.57 & 96.76 \\
			9 & 6.05 & 5.19 & 6.06 & 5.17 & \textbf{6.39} & 5.21 & 6.06 & 5.19  & 5.78 & 5.16 \\
			10 & \textbf{73.62} & 17.48 & \textbf{71.17} & 15.51 & \textbf{83.22} & 21.06 & \textbf{73.05 }& 35.57  & \textbf{64.30} & 18.87 \\
			11 & \textbf{99.15} & 77.51 & \textbf{99.15} & 76.83 & \textbf{99.12 }& 80.43 & \textbf{98.66} & 80.51  & \textbf{98.84} & 76.08 \\
			12 & 99.10 & 98.20 & \textbf{99.10} & 97.89 & \textbf{99.12} & 95.87 & \textbf{98.98} & 96.63  & 99.06 & 98.11 \\
			13 & \textbf{95.02} & 94.01 & \textbf{94.98} & 93.66 & \textbf{95.14} & 93.21 & \textbf{94.96} & 93.48  & 95.02 & 94.07 \\
			14 & 99.86 & 99.97 & 99.86 & 99.97 & 99.83 & 99.96 & 99.79 & 99.97  & 99.83 & 99.96 \\
			15 & \textbf{6.71} & 5.39 & \textbf{6.53} & 5.35 & \textbf{7.21} & 5.37 & \textbf{7.27 }& 5.36  & \textbf{7.43} & 5.48 \\
			16 & \textbf{73.72} & 13.43 & \textbf{75.15 }& 12.46 & \textbf{82.63} & 16.99 & \textbf{63.69} & 21.10  & \textbf{56.56} & 13.74 \\
			17 & \textbf{96.04} & 91.53 & 96.05 & 96.02 & 96.11 & 95.46 & 95.91 & 96.14  & \textbf{95.58} & 91.05 \\
			18 & 94.17 & 93.76 & 94.09 & 93.78 & 93.91 & 93.70 & 93.81 & 93.90  & 93.81 & 93.70 \\
			19 & \textbf{99.02} & 25.13 & \textbf{99.06} & 24.40 & \textbf{99.04} & 24.10 & \textbf{93.50} & 23.39  & \textbf{96.33} & 24.43 \\
			20 & \textbf{86.30} & 48.05 & \textbf{89.78} & 46.87 & \textbf{91.81} & 71.11 & \textbf{84.12} & 47.92  & \textbf{80.92} & 45.59 \\
			\midrule
			Average & \textbf{80.75} & 64.90 & \textbf{80.49} & 64.36 & \textbf{82.06} & 65.96 & \textbf{78.30} & 66.25  & \textbf{77.55} &  64.56\\
			\bottomrule
		\end{tabular}
	\label{table:tep}
\end{table*}
\subsubsection{TEP Dataset}
We report the FDR for all $20$ fault cases with a set FAR of $5\%$. For HAI dataset, we use both ROC
(Receiver Operating Characteristic) and PR (Precision-Recall) curves to evaluate the performance. AUC (Area Under
Curve) for the ROC curve and Average Precision (AP) are calculated as the overall performance measure. For
experiments presented in this section, the same set of hyperparemeters for each base model is used for the
proposed MAD task and the traditional NSP task, in order to make the comparison completely fair.

Results for TEP dataset are summarized in Table~\ref{table:tep} and grouped by base models. We
compare the fault detection results using our proposed MAD formulation and the traditional NSP formulation. The FAR
for the normal case is set to be 5\%, and the corresponding FDRs in percentage are presented in this table. It can be seen that with the
exact same base model and hyperparameters, our MAD approach significantly outperforms the NSP approach in many
cases (e.g., Fault \# 5, 10, 11, 16, 19, 20). Furthermore, for none of the faults and any base model, MAD
under-performs NSP by more than 1\% FDR. These results show that for an anomaly detection problem, one can
achieve better results from the same base model, by simply switching to our MAD framework for training and
inference.

\subsubsection{HAI Dataset}
Results for HAI dataset are reported in Table~\ref{table:hai} and also grouped by base models. We
compare the anomaly detection results using our proposed MAD formulation, its corresponding Fast-MAD 
evaluations, and the traditional NSP formulation. Note that for Fast-MAD we do not need to train a new model, but simply use the same MAD models with masks applied to only on the last step for anomaly detection.

\begin{table}[ht]
	\centering
	\caption{HAI dataset anomaly detection results. We report ROC AUC and Average Precision. For the same base model, higher values are in bold. We also list the average inference time for running the entire test set 5 times on the same hardware in seconds.}
		\begin{tabular}{c|c|c|c|c}
			\toprule
			Base Model & Method & ROC AUC & AP & Inference Time\\
			\midrule
			\multirow{3}{*}{Transformer} & MAD & \textbf{0.8032} & \textbf{0.4542} & 325.31s \\ 
			& Fast-MAD & 0.7916 & 0.3948 & 28.83s \\
			& NSP & 0.7839 & 0.3617 & 28.66s \\
			\midrule
			\multirow{3}{*}{\shortstack{Transformer-\\Encoder}} & MAD & \textbf{0.8160} & \textbf{0.5153} & 135.72s\\
			& Fast-MAD & 0.7973 & 0.3991 & 20.42s \\ 
			& NSP & 0.7411 & 0.3253 & 20.76s \\ 
			\midrule
			\multirow{3}{*}{LSTM} & MAD & \textbf{0.7715} & \textbf{0.4859} & 25.76s \\ 
			& Fast-MAD & 0.7432 & 0.3313 & 14.41s \\
			& NSP & 0.7441 & 0.2480 & 15.26s \\ 
			\midrule
			\multirow{3}{*}{TCN} & MAD & \textbf{0.7686} & \textbf{0.4580} & 27.48s \\ 
			& Fast-MAD & 0.7539 & 0.3508 & 14.46s \\ 
			& NSP & 0.7080 & 0.2389 & 15.11s \\ 
			\midrule
			\multirow{3}{*}{FNet} & MAD & \textbf{0.8112} & \textbf{0.5162} & 106.28s \\ 
			& Fast-MAD & 0.7949 & 0.3307 & 18.69s \\ 
			& NSP & 0.7754 & 0.3909 & 18.80s \\
			\bottomrule
		\end{tabular}
	\label{table:hai}
\end{table}

We report both the area under curve for ROC curves and the average precision (AP). The higher values for a given base model
are presented in bold. 
These results again show that for this multivariate industrial time series
anomaly detection problem, better results can be easily achieved by simply switching to our proposed MAD task from the
traditional NSP task for any given base model. 
Furthermore, the Fast-MAD results are in general unsurprisingly worse than MAD results, but still better or at least as good as NSP results. We also listed the average wall-clock time for testing the entire dataset on the same hardware in Table~\ref{table:hai}. Please note that these wall-clock values are not meant to provide a strictly quantitative comparison between different methods, because in reality there are many factors that can affect the elapsed time of running a program. Nevertheless, the point is that for the same base algorithm, Fast-MAD is almost exactly as fast as traditional NSP methods.


The improvements of MAD over NSP across the board can be explained by viewing MAD as a more generalized version of NSP, in the sense that the latter is similar
to only masking the last time step (Fast-MAD). Since Fast-MAD is performing at a better to similar level to NSP, and MAD is strictly 
better than Fast-MAD in terms of accuracy, MAD is therefore able to outperform NSP significantly. The comparable 
speed from Fast-MAD further solidifies the point that the proposed MAD formulation could replace NSP formulation in 
almost all real-world industrial settings. 
Even when computational
resources are limited, the proposed MAD-trained model can be easily adapted at inference time to run in Fast-MAD mode, because the systems that can already run NSP models should be able to run Fast-MAD models with the same hardware at the same speed. Furthermore, Fast-MAD inference would also not cause any delay in raising alarms compared to NSP inference, since they are both predicting the last step in the sequences.

\subsection{Ablation Studies}
In this subsection, we perform experiments over a number of factors in MAD to better understand the anomaly detection performances.
One major difference between MAD and BERT~\cite{devlin_bert_2019} masking procedures is that for language models, there can be a dedicated discrete mask token to indicate that an input has been masked, but in industrial time series data where many sensor data are not categorical but continuous, such discrete mask token does not exist. This is the reason that we used all random masks in MAD for results presented in Section~\ref{sec:exp:results}.

The following is an ablation study to evaluate the effect of different masking procedures for MAD. In this study, we 
keep the overall mask probability to be $0.15$. 
 Here, we have three masking rates for an input once it is chosen to be masked based on the aforementioned overall 
 probability: $p_{RND}$ is the probability that an input is replaced with a random number, $p_{SAME}$ is the 
 probability that an input is kept the same, and $p_{ZERO}$ is the probability that an input is replaced with a zero. 
 Furthermore, $p_{RND}+p_{ZERO}+p_{SAME}=1$. The default setting for the experiments in 
 Section~\ref{sec:exp:results} is that $p_{RND}=1$ and $p_{ZERO}=p_{SAME}=0$. That is to say, $100\%$ of the 
 $15\%$ selected inputs are masked with random, while none are masked with zero or kept the same.

\begin{table}[ht]
	\centering
	\caption{HAI dataset ablation study over different masking procedures. ROC AUC and AP are reported. The base model is Transformer-Encoder and kept the same across different masking strategies. If `Step' is true, then entire time steps are masked across all channels, otherwise only a subset of channels may be masked at a step.}
	\begin{tabular}{ccc|c|cc}
		\toprule
		\multicolumn{3}{c|}{Masking Rates} & \multirow{2}{*}{Step} & \multirow{2}{*}{ROC AUC} & \multirow{2}{*}{AP} \\
		$p_{RND}$ & $p_{SAME}$ & $p_{ZERO}$ &  &  &  \\
		\midrule
		100\% & 0\% & 0\% & True & 0.8160 & 0.5153 \\
		10\% & 10\% & 80\% & True & 0.7629 & 0.4809 \\
		0\% & 0\% & 100\% & True & 0.7331 & 0.2813 \\
		20\% & 0\% & 80\% & True & 0.7680 & 0.2484 \\
		0\% & 20\% & 80\% & True & 0.7169 & 0.1787 \\
		80\% & 20\% & 0\% & True & 0.8021 & 0.4955 \\
		\midrule
		100\% & 0\% & 0\% & False & 0.8085 & 0.4293 \\
		\bottomrule
	\end{tabular}
	\label{table:mask_prob}
\end{table}

In Table~\ref{table:mask_prob} we report the results of different masking rates on the same Transformer-Encoder 
model used in Section~\ref{sec:exp:results}. We also explored a different masking strategy, i.e., instead of  masking 
all input dimensions for selected time steps (\textit{Step=True} in Table~\ref{table:mask_prob}), we can randomly mask 
$15\%$ of the inputs (i.e. only some dimensions for certain time steps are masked, \textit{Step=False} in 
Table~\ref{table:mask_prob}). The first row in this table corresponds to the default strategy we used in 
Section~\ref{sec:exp:results}, i.e., all of 15\% selected time steps are masked 100\% with random values, and it 
gives the best overall performance. We believe for continuous time series, replacing with zeros is counterproductive 
because a $0$ value has physical meanings, so it is inappropriate as a mask token. Instead, the 
best strategy is to just replace masked inputs with random values within the ranges of the input.

\section{Conclusion}
In this paper, we proposed and validated a self-supervised learning task, \textbf{M}asked \textbf{A}nomaly 
\textbf{D}etection (\textbf{MAD}), for multivariate time series anomaly detection. 
The main research angle of this paper is not to find an optimal set of hyperparameters for a particular dataset, but 
rather to
show that in general MAD tasks can outperform traditional next step prediction (NSP) tasks in anomaly detection, even when using exactly the same base model, and they can run as fast as NSP models during testing if needed. 
We also investigated different design choices for our MAD 
formulation, including different evaluation techniques and masking procedures, highlighting  best practices for using 
this proposed method. 
Since our proposed MAD task is not restricted to a particular neural network 
architecture,  existing applications using NSP can easily switch to MAD without changing the base 
model or hardware and gain performance improvements. Therefore, we believe our proposed MAD task has 
significant impacts on real-world industrial anomaly detection applications. 

\section*{Acknowledgment}
This material is based upon work supported by the Department of Energy, National Energy Technology Laboratory under Award Number DE-FE0031763.\footnote{Disclaimer:  This report was prepared as an account of work sponsored by an agency of the United States Government.  Neither the United States Government nor any agency thereof, nor any of their employees, makes any warranty, express or implied, or assumes any legal liability or responsibility for the accuracy, completeness, or usefulness of any information, apparatus, product, or process disclosed, or represents that its use would not infringe privately owned rights.  Reference herein to any specific commercial product, process, or service by trade name, trademark, manufacturer, or otherwise does not necessarily constitute or imply its endorsement, recommendation, or favoring by the United States Government or any agency thereof.  The views and opinions of authors expressed herein do not necessarily state or reflect those of the United States Government or any agency thereof.}

\bibliographystyle{IEEEtran}
\bibliography{arxiv}

\end{document}